\pdfoutput=1
\documentclass[11pt]{article}

\usepackage[preprint]{acl}

\usepackage{times}
\usepackage{latexsym}

\usepackage{amsmath}
\usepackage{mathrsfs}
\usepackage{amsfonts}

\usepackage[T1]{fontenc}
\usepackage[utf8]{inputenc}

\usepackage{microtype}

\usepackage{inconsolata}

\usepackage{graphicx}

\usepackage[cjk]{kotex}
\usepackage{tabularx}
\usepackage{arydshln}

\newcommand*{\jp}[1]{{\color{black}#1}}

\title{Enhancing Korean Dependency Parsing with Morphosyntactic Features}

\author{
Jungyeul Park$^{1}$~~~ Yige Chen$^{2}$~~~ Kyuwon Kim$^{3}$~~~ KyungTae Lim$^{4}$~~~ Chulwoo Park$^{5}$\\
$^{1}$The University of British Columbia, Canada\\
$^{2}$The Chinese University of Hong Kong, Hong Kong\\
$^{3}$Seoul National University, South Korea\\
$^{4}$KAIST, South Korea~~~
$^{5}$Anyang University, South Korea\\
\url{https://github.com/jungyeul/k-unidive}
}

\begin{document}
\maketitle

\begin{abstract}
This paper introduces UniDive for Korean, an integrated framework that bridges Universal Dependencies (UD) and Universal Morphology (UniMorph) to enhance the representation and processing of Korean {morphosyntax}. Korean's rich inflectional morphology and flexible word order pose challenges for existing frameworks, which often treat morphology and syntax separately, leading to inconsistencies in linguistic analysis. UniDive unifies syntactic and morphological annotations by preserving syntactic dependencies while incorporating UniMorph-derived features, improving consistency in annotation. We construct an integrated dataset and apply it to dependency parsing, demonstrating that enriched morphosyntactic features enhance parsing accuracy, particularly in distinguishing grammatical relations influenced by morphology. Our experiments, conducted with both encoder-only and decoder-only models, confirm that explicit morphological information contributes to more accurate syntactic analysis.
% UniDive could offer a scalable approach to other morphologically rich languages by harmonizing morphological and syntactic structures. By integrating UD and UniMorph, this framework advances linguistic resource development and computational efficiency, facilitating more precise analysis of complex morphosyntactic interactions for multilingual NLP.
\end{abstract}

\section{Introduction}

Modeling morphosyntactic structures in agglutinative languages such as Korean presents significant challenges due to their rich inflectional morphology, non-trivial word segmentation, and complex syntactic constructions. 
Unlike isolating languages, where words and morphemes align clearly with syntactic units, Korean's extensive use of suffixes to express grammatical relations complicates traditional approaches to syntactic annotation and parsing. As a result, computational frameworks must go beyond conventional word-based representations to effectively capture both morphological and syntactic properties.

The Universal Dependencies (UD) framework \citep{de-marneffe-etal-2021-universal} has emerged as a widely adopted standard for syntactic annotation, facilitating cross-linguistic comparisons and multilingual NLP. However, its treatment of morphological information remains limited, especially in agglutinative languages like Korean, where morphosyntactic dependencies are deeply intertwined. The UniMorph project \citep{batsuren-etal-2022-unimorph} complements UD by providing a detailed morphological schema for inflectional paradigms but does not encode syntactic dependencies. The lack of integration between these two frameworks leads to inconsistencies, particularly when representing {functional} morphemes, argument structures, and the boundary between morphology and syntax.
To address these issues, we explore the integration of UniMorph and UD under the UniDive framework, which aims to harmonize morphological and syntactic representations in a more systematic way. UniDive proposes a morphosyntactic data structure where {lexical} words are represented as separate nodes in a dependency graph, while {functional} words and morphemes contribute feature-based information rather than being treated as independent syntactic units. This approach not only reduces segmentation inconsistencies but also enables better handling of non-concatenative morphology, periphrastic expressions, and argument structures encapsulated within words.

Recent efforts in Korean linguistic resources, such as K-UD \citep{kim-etal-2024-kud} and K-UniMorph \citep{jo-etal-2023-k}, have attempted to address these challenges. K-UD provides a syntactically annotated corpus following the UD framework \citep{noh-etal-2018-enhancing,chun-EtAl:2018:LREC}, while K-UniMorph offers a morphological dataset aligned with UniMorph's schema. However, these resources were developed independently, leading to gaps and inconsistencies in transitioning between morphological and syntactic representations.

This paper introduces a UniDive dataset for Korean, designed to enhance the representation of Korean morphosyntactic structures by integrating Google's Korean UD \citep{mcdonald-etal-2013-universal} within the UniDive framework. We present its practical implementation, demonstrating how this approach improves syntactic parsing, enhances morphosyntactic analysis, and benefits downstream NLP tasks. By adopting this unified representation, we aim to contribute to the development of linguistic resources for Korean and other agglutinative languages while advancing computational methodologies for processing morphologically rich languages.

\section{Related Work}

The Universal Dependencies (UD) framework has established itself as a leading standard for syntactic annotation across a wide variety of languages. Its primary goal is to provide a unified, language-agnostic representation of syntactic structures, enabling cross-linguistic comparisons and facilitating multilingual natural language processing (NLP) tasks. However, while UD has proven effective for many languages, its ability to represent the morphosyntactic complexity of agglutinative languages such as Korean remains limited. The representation of fine-grained morphological details, which are essential for accurate syntactic parsing in Korean, often lacks depth in UD's annotation scheme \citep{chen-etal-2022-yet}.

The UniMorph project complements UD by focusing on the morphological aspect of linguistic data. UniMorph standardizes the representation of inflectional paradigms across languages, offering a rich resource for morphological analysis. While UniMorph provides detailed morphological annotations, it does not explicitly encode syntactic dependencies. This separation between morphology and syntax poses challenges for languages like Korean, where morphological and syntactic properties are {inherently associated}. For instance, Korean's extensive use of suffixes to denote case, tense, and aspect requires a more integrated approach to handle morphosyntactic interactions effectively.

Prior efforts to address these limitations include language-specific adaptations of UD and UniMorph. Projects such as K-UD \citep{kim-etal-2024-kud} and K-UniMorph \citep{jo-etal-2023-k} have made significant {contributions} in applying these frameworks to Korean. K-UD offers syntactic annotations within the UD framework, while K-UniMorph provides a rich set of morphological annotations aligned with the UniMorph schema. Despite these advancements, the lack of integration between K-UD and K-UniMorph limits their utility for tasks requiring a unified morphosyntactic representation.

UniDive introduces a novel approach to bridging this gap by {integrating} UD and UniMorph into a single, cohesive framework. By aligning morphological and syntactic representations, UniDive ensures consistency and compatibility, enabling a more comprehensive analysis of morphosyntactic structures. This approach is particularly beneficial for morphologically rich languages like Korean, where existing frameworks often fail to capture the intricate relationships between morphology and syntax.
Additionally, the UniDive framework extends the applicability of UD and UniMorph to other morphologically rich languages, addressing challenges such as non-concatenative morphology, complex agreement patterns, and free word order. By providing a unified structure, UniDive facilitates the development of linguistic resources that are both theoretically robust and computationally efficient.
Our work builds on these prior efforts by introducing an integrated dataset for Korean that aligns K-UD and K-UniMorph within the UniDive framework.

\section{Morphosyntactic Features}

In Universal Dependencies, morphosyntactic features provide detailed grammatical information about words, {which defines} their tense, aspect, mood, case, number, person, politeness level, and more. These features contribute to accurate syntactic parsing, morphological analysis, and cross-linguistic comparison. Below is an introduction to key UD morphosyntactic features:

\textsc{Case} marks the grammatical role of a noun phrase in a sentence, such as ablative, accusative, conjunctive, dative, disjunctive, genitive, instrumental, locative, or nominative.
\textsc{Evident} refers to evidentiality, distinguishing whether the speaker has firsthand knowledge, inferred knowledge, non-firsthand knowledge, or reported information.
\textsc{Mood} expresses the attitude of the speaker toward the action, including conditional, general conditional, general-potential conditional, potential conditional, desiderative, imperative, indicative, interrogative, necessity, optative, and potential moods.
\textsc{NumType} identifies numerals and their types, such as cardinal numbers.
\textsc{Person} distinguishes between singular and plural forms, with plural being explicitly marked.
Person indicates who is performing the action, whether first person, second person, or third person.
\textsc{Person[psor]} specifies the possessor's person in possessive structures, with distinctions for first person, second person, and third person.
\textsc{Polite} reflects the level of politeness or formality in speech, with values for elevated politeness, formal speech, and humble speech.
\textsc{PronType} categorizes pronouns into different types, such as articles, demonstratives, indefinite pronouns, interrogative pronouns, personal pronouns, and reciprocal pronouns.
\textsc{Tense} conveys time reference, distinguishing between present and other tenses.
\textsc{VerbForm} describes how a verb is used, such as converb, finite verb, participle, or verbal noun.
\textsc{Voice} indicates the relationship between the verb's subject and the action, including causative, causative-passive, passive, reciprocal, and reflexive constructions.
% \end{itemize}

Table~\ref{ms-table} shows the morphosyntactic features used in UD and their corresponding annotations in Korean. These features capture key grammatical distinctions across languages and help morphological tagging, syntactic analysis, and linguistic resource development. They include \textsc{Aspect}, which differentiates habitual, perfective, and progressive actions; \textsc{Case}, which marks the grammatical roles of nouns; \textsc{Evidentiality}, which indicates the source of knowledge; and \textsc{Mood}, which reflects the speaker's attitude or modality. Additionally, \textsc{Tense} and \textsc{VerbForm} distinguish between different time references and verbal inflections, while \textsc{Voice} accounts for relationships between subjects and actions. These morphosyntactic annotations provide a structured framework for analyzing linguistic variations and ensuring cross-linguistic compatibility in natural language processing tasks.

\begin{table*}[!ht]
    \centering
\tiny{
\begin{tabularx}{\textwidth}{ll X X}
% &  UD Annotation & Korean Examples \\

\hline
\textsc{Aspect} 
& Habitual (\texttt{Aspect=Hab}) & 하곤 했다 \textit{hagon haetda} & (`used to' \textit{\color{gray}repeated or habitual actions in the past}) \\
& Perfective (\texttt{Aspect=Perf}) & 했다 \textit{haetda} & (`has done' \textit{\color{gray}completed actions or results}) \\
& Progressive (\texttt{Aspect=Prog}) & 하고 있다 \textit{hago itda} & (`is doing' \textit{\color{gray}ongoing or continuous actions}) \\
\hdashline

\textsc{Case} 
& Ablative (\texttt{Case=Abl}) & 에서 \textit{eseo}, 부터 \textit{buteo} & (`from' \textit{\color{gray}a place or time point}) \\
& Accusative (\texttt{Case=Acc}) & 을 \textit{eul}, 를 \textit{reul} & (`marks direct object') \\
& Conjunctive (\texttt{Case=Conj}) & 와 \textit{wa}, 과 \textit{gwa} & (`and') \\
& Dative (\texttt{Case=Dat}) & 에게 \textit{ege}, 한테 \textit{hante}, 께 \textit{kke} & (`to, for' \textit{\color{gray}recipient or indirect object}) \\
& Disjunctive (\texttt{Case=Disj}) & 이나 \textit{ina}, 나 \textit{na} & (`or') \\
& Genitive (\texttt{Case=Gen}) & 의 \textit{ui} & (`possessive, of, 's') \\
& Instrumental (\texttt{Case=Ins}) & 로 \textit{ro} & (`by, using, with') \\
& Locative (\texttt{Case=Loc}) & 에 \textit{e} & (`at, in, on' \textit{\color{gray}a location}) \\
& Nominative (\texttt{Case=Nom}) & 이 \textit{i}, 가 \textit{ga}, 은 \textit{eun}, 는 \textit{neun} & (`subject marker') \\
\hdashline

\textsc{Evident} 
& Firsthand (\texttt{Evident=Fh}) & 내가 봤어 \textit{naega bwasseo} & (`I saw it myself' \textit{\color{gray}directly witnessed event}) \\
& Inferential (\texttt{Evident=Infer}) & 비가 올 것 같다 \textit{biga ol geot gata} & (`It looks like it will rain' \textit{\color{gray}deduced or inferred from evidence}) \\
& Non-firsthand (\texttt{Evident=Nfh}) & 들었어 \textit{deureosseo} & (`I heard about it' \textit{\color{gray}information learned from others}) \\
& Reported (\texttt{Evident=Rep}) & 비가 왔다고 해 \textit{biga watdago hae} & (`They say it rained' \textit{\color{gray}reported or quoted speech}) \\
\hdashline

\textsc{Mood}
& Conditional (\texttt{Mood=Cnd}) & 비가 오면 갈게 \textit{biga omyeon galge} & (`I will go if it rains' \textit{\color{gray}hypothetical condition}) \\
& General Conditional (\texttt{Mood=CndGen}) & 사람이면 누구나 실수한다 \textit{saramimyeon nuguna silsuhanda} & (`Anyone can make mistakes' \textit{\color{gray}general truth under a condition}) \\
& Potential Conditional (\texttt{Mood=CndPot}) & 시간이 있으면 도울 수 있어 \textit{sigani isseumyeon doul su isseo} & (`I can help if I have time' \textit{\color{gray}potential action under a condition}) \\
& General-Potential Conditional (\texttt{Mood=CndGenPot}) & 건강하면 오래 산다 \textit{geonganghamyeon orae sanda} & (`If you are healthy, you live long' \textit{\color{gray}general possibility based on a condition}) \\
& Desiderative (\texttt{Mood=Des}) & 가고 싶다 \textit{gago sipda} & (`I want to go' \textit{\color{gray}expressing desire}) \\
& Imperative (\texttt{Mood=Imp}) & 조용히 해! \textit{joyonghi hae} & (`Be quiet!' \textit{\color{gray}giving a command}) \\
& Indicative (\texttt{Mood=Ind}) & 나는 학교에 간다 \textit{naneun hakgyoe ganda} & (`I go to school' \textit{\color{gray}neutral statement of fact}) \\
& Interrogative (\texttt{Mood=Int}) & 어디에 가니? \textit{eodie gani?} & (`Where are you going?' \textit{\color{gray}asking a question}) \\
& Necessitative (\texttt{Mood=Nec}) & 가야 한다 \textit{gaya handa} & (`I must go' \textit{\color{gray}expressing necessity}) \\
& Optative (\texttt{Mood=Opt}) & 행복하길 바란다 \textit{haengbokhagil baranda} & (`I hope you will be happy' \textit{\color{gray}expressing a wish}) \\
& Potential (\texttt{Mood=Pot}) & 할 수 있다 \textit{hal su itda} & (`I can do it' \textit{\color{gray}expressing ability or possibility}) \\
\hdashline

\textsc{NumType}
& Cardinal Number (\texttt{NumType=Card}) & 세 개 \textit{se gae}, 다섯 명 \textit{daseot myeong} & (`three things', `five people' \textit{\color{gray}counting objects or people}) \\
& Plural (\texttt{Number=Plur}) & 학생들 \textit{haksaengdeul} & (`students' \textit{\color{gray}plural marker}) \\
\hdashline

\textsc{Person}
& First Person (\texttt{Person=1}) & 나는 \textit{naneun}, 우리가 \textit{uriga} & (`I', `we' \textit{\color{gray}speaker including self}) \\
& Second Person (\texttt{Person=2}) & 너는 \textit{neoneun}, 당신이 \textit{dangsin-i} & (`you' \textit{\color{gray}listener/addressee}) \\
& Third Person (\texttt{Person=3}) & 그는 \textit{geuneun}, 그들은 \textit{geudeureun} & (`he', `they' \textit{\color{gray}third party reference}) \\
\hdashline

\textsc{Person[psor]}
& First Person Possessive (\texttt{Person[psor]=1}) & 내 책 \textit{nae chaek}, 우리 집 \textit{uri jip} & (`my book', `our house' \textit{\color{gray}first-person possession}) \\
& Second Person Possessive (\texttt{Person[psor]=2}) & 네 가방 \textit{ne gabang} & (`your bag' \textit{\color{gray}second-person possession}) \\
& Third Person Possessive (\texttt{Person[psor]=3}) & 그의 차 \textit{geuui cha}, 그들의 의견 \textit{geudeurui uigyeon} & (`his car', `their opinion' \textit{\color{gray}third-person possession}) \\
\hdashline

\textsc{Polite}
& Elevated (\texttt{Polite=Elev}) & 선생님께서 오십니다 \textit{seonsaengnimkkeseo osimnida} & (`The teacher is coming' \textit{\color{gray}honorific subject marking}) \\
& Formal (\texttt{Polite=Form}) & 갑니다 \textit{gamnida}, 합니다 \textit{hamnida} & (`go', `do' \textit{\color{gray}formal speech level}) \\
& Humble (\texttt{Polite=Humb}) & 드리겠습니다 \textit{deurigessseubnida} & (`I will give (to you)' \textit{\color{gray}humble expression of giving}) \\
\hdashline

\textsc{PronType}
& Article (\texttt{PronType=Art}) & 그 책 \textit{geu chaek} & (`the book' \textit{\color{gray}definite article-like determiner}) \\
& Demonstrative (\texttt{PronType=Dem}) & 이 사람 \textit{i saram}, 저 집 \textit{jeo jip} & (`this person', `that house' \textit{\color{gray}indicating specific reference}) \\
& Indefinite (\texttt{PronType=Ind}) & 어떤 사람 \textit{eotteon saram}, 아무도 \textit{amudo} & (`some person', `nobody' \textit{\color{gray}unspecified reference}) \\
& Interrogative (\texttt{PronType=Int}) & 누구 \textit{nugu}, 무엇 \textit{mueot} & (`who', `what' \textit{\color{gray}used in questions}) \\
& Personal (\texttt{PronType=Prs}) & 나는 \textit{naneun}, 그들은 \textit{geudeureun} & (`I', `they' \textit{\color{gray}referencing persons}) \\
& Reciprocal (\texttt{PronType=Rcp}) & 서로 \textit{seoro} & (`each other' \textit{\color{gray}mutual action reference}) \\
\hdashline

\textsc{Tense}
& Present (\texttt{Tense=Pres}) & 먹는다 \textit{meokneunda}, 공부한다 \textit{gongbuhanda} & (`eats', `studies' \textit{\color{gray}present or habitual action}) \\
& Past (\texttt{Tense=Past}) & 먹었다 \textit{meogeotda}, 공부했다 \textit{gongbuhaetda} & (`ate', `studied' \textit{\color{gray}completed action in the past}) \\
\hdashline

\textsc{VerbForm}
& Converb (\texttt{VerbForm=Conv}) & 먹고 \textit{meokgo}, 공부하며 \textit{gongbuhamyeo} & (`eating', `while studying' \textit{\color{gray}expressing simultaneous or sequential actions}) \\
& Finite (\texttt{VerbForm=Fin}) & 먹는다 \textit{meokneunda}, 공부한다 \textit{gongbuhanda} & (`eats', `studies' \textit{\color{gray}main finite verb in a clause}) \\
& Participle (\texttt{VerbForm=Part}) & 먹은 \textit{meogeun}, 공부한 \textit{gongbuhan} & (`eaten', `studied' \textit{\color{gray}adjectival participle modifying a noun}) \\
& Verbal Noun (\texttt{VerbForm=Vnoun}) & 먹기 \textit{meokgi}, 공부하기 \textit{gongbuhagi} & (`eating', `studying' \textit{\color{gray}verbal noun used as a subject or object}) \\
\hdashline

\textsc{Voice}
& Causative (\texttt{Voice=Cau}) & 먹였다 \textit{meogyeotda}, 울렸다 \textit{ullyeotda} & (`made someone eat', `made someone cry' \textit{\color{gray}causative action forced by subject}) \\
& Causative-Passive (\texttt{Voice=CauPass}) & 보였다 \textit{boyeotda}, 들렸다 \textit{deullyeotda} & (`was shown', `was heard' \textit{\color{gray}causative-passive meaning}) \\
& Passive (\texttt{Voice=Pass}) & 먹혔다 \textit{meokhyeotda}, 잡혔다 \textit{japhyeotda} & (`was eaten', `was caught' \textit{\color{gray}action performed on the subject}) \\
& Reciprocal (\texttt{Voice=Rcp}) & 만났다 \textit{mannatda}, 싸웠다 \textit{ssawotda} & (`met each other', `fought each other' \textit{\color{gray}mutual action between subjects}) \\
& Reflexive (\texttt{Voice=Rfl}) & 씻었다 \textit{ssiseotda}, 숨었다 \textit{sumeotda} & (`washed oneself', `hid oneself' \textit{\color{gray}subject performs action on itself}) \\
\hline 
\end{tabularx}
}
    \caption{Morphosyntactic features for Korean}
    \label{ms-table}
\end{table*}

The updated morphosyntactic features for Korean compared to K-UniMorph \citep{jo-etal-2023-k} introduce finer distinctions in case marking, evidentiality, interrogativity, mood, and voice, significantly enhancing the accuracy of linguistic annotation. In case marking, new categories such as comitative, vocative, comparative, and allative have been introduced to better capture distinctions in accompaniment, direct address, comparison, and movement towards a location. These refinements provide a more precise representation of noun phrase functions that were previously conflated under broader categories.

Evidentiality and mood have also been significantly expanded. While previous versions {recognize} inferential evidentiality, the new system includes hearsay and non-firsthand evidentiality, which are essential for distinguishing reported speech and indirectly obtained knowledge in Korean. Mood distinctions have been extended to include realis and irrealis, marking factual versus hypothetical statements, as well as purposive and obligative moods, which indicate intent and necessity. These refinements improve the representation of modality, particularly in formal and polite speech.

The voice system has {also} been expanded with the addition of causative-passive, reciprocal, and reflexive voices, clarifying agent-patient relationships in complex predicates. The inclusion of explicit declarative and interrogative markers further distinguishes statement and question structures. Overall, these enhancements ensure a more comprehensive and language-specific analysis of Korean grammar, improving both syntactic parsing and cross-linguistic compatibility.

\section{Parsing with Generative LLMs}
BERT models have been widely employed in parsing tasks due to their ability to capture bidirectional contextual information, which has proven advantageous over decoder-only architectures \citep{zeman-etal-2018-conll, lim-etal-2020-semi-supervised}. However, with the rapid advancement of generative language models, recent research has explored the feasibility of replacing {encoder-only} models with generative architectures \citep{lin-etal-2023-chatgpt, xie-etal-2023-empirical, tian-etal-2024-large}. Notably, \citet{tian-etal-2024-large} {propose} a constituency parsing approach utilizing a {decoder-only} generative {language} model, structuring the parsing process into three distinct stages. While their model attained approximately 87\% of the performance achieved by {encoder-only} models, the primary limitation {is} attributed to the generative model's difficulty in accurately segmenting chunks within input sentences.

The Llama3.2-3B\footnote{\url{huggingface.co/meta-llama/Llama-3.2-3B}} model, a well-known generative language model from the Llama family, serves as the baseline for evaluation. However, the base Llama3.2 model demonstrates extremely limited proficiency in Korean, making effective training infeasible. To address this limitation, the Bllossom-3B\footnote{\url{huggingface.co/Bllossom/llama-3.2-Korean-Bllossom-3B}} model~\citep{choi-etal-2024-optimizing}, an enhanced version of Llama3.2-3B optimized for Korean, is employed. Further fine-tuning is performed using the proposed Korean UniDive dataset for dependency parsing. This process requires converting the Korean UniDive data into an {instruction tuning} (IT) format, following the IT training data structure proposed by Stanford Alpaca~\citep{taori-etal-2023-stanford-alpaca}.

This instruction tuning dataset comprises three components: the user's intent (\texttt{Instruction}), the provided input data (\texttt{Input}), and the expected output data (\texttt{Output}). Figure~\ref{instruction-data-dp} illustrates the process of transforming Korean UniDive training data into an instruction tuning format. Consequently, the entire training set of the Korean UniDive dataset is converted into IT-formatted data for model training.
The instruction tuning process follows a next-word prediction paradigm, akin to causal language modeling (CLM) pretraining commonly used for language models. The key distinction lies in the loss computation, where feedback is applied exclusively to the \texttt{Output} portion that the model is required to generate, while the \texttt{Instruction} and \texttt{Input} components remain unpenalized. This process can be formally expressed using the following loss function:
\begin{equation}
    L_{\!it}(\theta)\!=\!\mathbb{E}_{x \sim \mathscr{D}_{\!IT}}\!\left\{\! -\!\sum_{\!i\in out\!}{logP(x_{i}|x_{\!<i};\theta)}\! \right\}
    \label{eq-sft}
\end{equation}
 \noindent where \(\theta\) denotes the model parameters, and \(\mathscr{D}_{IT}\) refers to the dataset obtained by converting the previously proposed UniDive training data into an instruction tuning dataset. Each training sample can be represented as a token sequence \(x = (x_{0}, x_{1}, \ldots, x_{i})\), which tokenizes \texttt{instruction}, \texttt{input}, and \texttt{output}, as illustrated in Figure~\ref{instruction-data-dp}. After completing the IT stage, the model generates parsing results for the given \texttt{input}. For evaluation, we {convert} all samples in the original test data into the IT format, following the \texttt{Instruction}, \texttt{Input}, \texttt{Output} example in Figure~\ref{instruction-data-dp}, and then provided only \texttt{Instruction} and \texttt{Input} to the model to predict the \texttt{Output}. Detailed information on the hyperparameters used for training, computational resources, and training duration can be found in Appendix~\ref{appendix:exp-settings}.

\begin{figure}[!ht]
\centering
\resizebox{.5\textwidth}{!}{
{
\footnotesize{
\begin{tabular}{llllllll}
\multicolumn{8}{l}{\textbf{[\texttt{instruction}]}: 아래의 문장을 의존구조문법에 맞게 분석해줘} \\
\multicolumn{8}{l}{{(`Please parse the following sentence as a formal dependency parsing')}} \\
~\\
\multicolumn{8}{l}{\textbf{[\texttt{input}]:}} \\
1 & 학교 & 학교 & NOUN & NNG & \_ & \texttt{head} & \texttt{rel} \\
2 & 분위기나 & 분위기+나 & NOUN & NNG+JC & Case=Disj & \texttt{head} & \texttt{rel} \\
3 & 경관이 & 경관+이 & NOUN & NNG+JKS & Case=Nom & \texttt{head} & \texttt{rel} \\
4 & 굉장히 & 굉장히 & ADV & MAG & \_ & \texttt{head} & \texttt{rel} \\
5 & 좋다 & 좋+다 & ADJ & VA+EF & Mood=Ind & \texttt{head} & \texttt{rel} \\
6 & . & . & PUNCT & SF & \_ & \texttt{head} & \texttt{rel} \\
~\\
\multicolumn{8}{l}{\textbf{[\texttt{output}]:}} \\
1 & 학교 & 학교 & NOUN & NNG & \_ & \textbf{5} & \textbf{nsubj} \\
2 & 분위기나 & 분위기+나 & NOUN & NNG+JC & Case=Disj & \textbf{1} & \textbf{flat} \\
3 & 경관이 & 경관+이 & NOUN & NNG+JKS & Case=Nom & \textbf{1} & \textbf{conj} \\
4 & 굉장히 & 굉장히 & ADV & MAG & \_ & \textbf{5} & \textbf{advmod} \\
5 & 좋다 & 좋+다 & ADJ & VA+EF & Mood=Ind & \textbf{0} & \textbf{root} \\
6 & . & . & PUNCT & SF & \_ & \textbf{5} & \textbf{punct} \\
\multicolumn{8}{l}{{(`The school atmosphere and scenery are very good.')}} \\
% \hline
% \end{tabular}
\end{tabular}
}
}
}
\caption{Example of the instruction tuning data} \label{instruction-data-dp}
\end{figure}

\section{Experiments and Results}

\subsection{Korean GSD}

We refine {UD's} Korean GSD treebank for the Korean UniDive dataset by addressing inconsistencies in {part-of-speech} (POS) tagging and morphological analysis. By integrating UniMorph-derived features and applying systematic corrections, the dataset enhances syntactic parsing accuracy and morphosyntactic analysis, which are essential for improving dependency parsing and NLP applications that rely on precise linguistic annotations.

Each word in the Korean UniDive dataset is enriched with UniMorph-derived features, assigned using deterministic rules based on morphological patterns observed in surface forms. These features are integrated directly into the UD format, ensuring compatibility with existing UD tools. To maintain consistency and accuracy, we employ a two-step approach: (i) automatic annotation using established tagging tools, including POS {tagging} \citep{park-tyers:2019:LAW} and {Named Entity Recognition (NER)} \citep{chen-lim-park-2024-korean} applied to UD-annotated sentences, and (ii) manual verification by linguistic experts, who validate the assigned POS tags. This approach systematically refines the dataset and improves its reliability.

Each word in the Korean UniDive dataset is further analyzed at the morpheme level, enabling the extraction of morphosyntactic features based on both individual morphemes and their corresponding POS tags. These features are assigned using a set of detailed linguistic rules that account for Korean's agglutinative morphology, including suffix-based inflection, grammatical function markers, and verb-final {particles}. By leveraging fine-grained segmentation and feature extraction, we systematically capture case markers, tense and aspect distinctions, and honorifics, ensuring that morphosyntactic information is accurately preserved. This rule-based refinement not only improves alignment between morphological and syntactic annotations but also enhances consistency across dependency structures, making the dataset more robust for parsing and NLP applications.

\paragraph{POS misclassification}  
One of the most frequent errors observed in the {Korean GSD treebank is} the misclassification of proper nouns (NNP) as common nouns (NNG) in their {language-specific part-of-speech (XPOS)} labels. Additionally, the corresponding {universal part-of-speech} (UPOS) tags were often incorrectly assigned, failing to distinguish between proper nouns (PROPN) and common nouns (NOUN) at the universal level.

Beyond noun-related errors, verbs (VV) and adjectives (VA) were also frequently mislabeled. In many cases, verbs were incorrectly tagged as adjectives (ADJ), particularly when descriptive verbs were involved, as Korean adjectives (VA) function similarly to stative verbs. Likewise, verb-adjective misclassification affected UPOS, where Korean adjectives (VA) were incorrectly labeled as verbs (VERB), despite their distinct morphological and syntactic behaviors.

To correct these errors, we systematically verify the consistency between POS tagging and NER results for nouns while also applying morphosyntactic analysis for verbs and adjectives. If a token is identified as both a proper noun (NNP) and a named entity in NER but is incorrectly labeled as a common noun (NNG), we reclassify it as a proper noun (PROPN). Conversely, if a token is not recognized as a named entity but is incorrectly tagged as a proper noun (NNP), we adjust it to a common noun (NNG, NOUN).

For verbs and adjectives, we analyze inflectional patterns and conjugational endings to ensure correct classification. If a verb (VV) is incorrectly tagged as an adjective (VA), we examine morphological cues and predicate structure to determine whether the word exhibits verbal or adjectival properties. This approach ensures that adjectives (VA) are correctly assigned to ADJ and verbs (VV) to VERB, preserving the accuracy of the dependency structure.

Additionally, we observed inconsistencies in the annotation of inflected verb and adjective forms, where derivational suffixes led to misclassifications between adverbs (ADV), adjectives (ADJ), and verbs (VERB). In particular, verb-derived adjectives were frequently misclassified as adverbs (ADV) instead of adjectives (ADJ) due to their sentence context.
To address these inconsistencies, we reviewed and corrected POS information, including morphological analysis with explicit morpheme boundaries, following Sejong corpus guidelines to ensure that each word's annotation accurately reflected its true morphological category. However, these corrections were limited to POS tagging and morphological segmentation, with {no modification to} the dependency structure.

\paragraph{POS assignment based on syntactic roles}  
A significant source of error in the {Korean GSD treebank is} the misclassification of words due to an overemphasis on their syntactic {roles} within a given sentence rather than their inherent lexical {categories}. This issue led to frequent misannotations, particularly in cases where temporal nouns and other category-specific words were assigned adverbial roles based on their sentence context rather than their fundamental lexical identity.  
One prominent example is {가격에} \textit{gagyeog-e} (`price+\textsc{-e}'), which was originally labeled as ADV instead of its correct annotation as NOUN for NNG+JKB (common noun + adverbial postposition). This misclassification likely stemmed from the word's function as a modifier in specific contexts, leading to an erroneous assignment of an adverbial tag. However, based on the Sejong POS guidelines, such words should retain their noun classification regardless of their syntactic role in a particular sentence.
To address these issues, we systematically revised the affected annotations, ensuring that UPOS labels were determined based on the lexical properties of each word rather than their contextual function within a sentence. This correction improves consistency in POS tagging and aligns with best practices for Korean morphological annotation. 

\paragraph{Misclassification of XR as a noun fragment}
XR is defined as a root, which is the core part of a word that carries the essential meaning when analyzing words. However, in the Sejong Corpus, it is defined as a noun fragment that {does not function independently}.
For example, the root 민주 \textit{minju} can form a noun when combined with other affixes, such as in 민주화 \textit{minjuhwa} (`democratization') or 민주주의 \textit{minjujuui} (`democracy'). Only when it becomes a noun can it combine with derivational endings to be transformed into other parts of speech.

\paragraph{Complement markers in Korean grammar}
In Korean grammar, \textit{nominals} that appear before the verbs 되다 \textit{doeda} (`become') and 아니다 \textit{anida} (`not be') function as complements (predicate nominatives) rather than subjects. This is because 되다 \textit{doeda} and 아니다 \textit{anida} act as linking verbs (copulas) rather than action verbs.
Additionally, while these complements take the subject marker (-이/가, -\textit{i}/-\textit{ga}), they function as predicate complements rather than typical subjects in the sentence structure.
Therefore, their case marker should be annotated as JKC (predicate nominative marker) instead of JKS (nominative case marker).

We also revise the morphological analysis after manual verification, correcting cases where the GSD annotation and automatic annotation differ. For example, the analysis of 중일 as 중+이+ㄹ;VERB;NNB+VCP+ETM (bound noun + copula + adnominal ending) is corrected to 중일;PROPN;NNP \textit{jung-il} (`China and Japan').
By refining the UPOS and XPOS annotations, we have enhanced the reliability of {the GSD treebank} as a training resource for Korean syntactic parsing and morphological analysis, particularly for deriving morphosyntactic features from POS tags. Our systematic corrections in the {Korean GSD} dataset effectively address major inconsistencies. 
Table~\ref{pos-conversion-results} presents the top five UPOS and XPOS conversion results.

\begin{table*}[!ht]
\centering
\footnotesize{
\begin{tabular}{rcl cc | rcl cc} \hline 
\multicolumn{5}{c}{UPOS correction} & \multicolumn{5}{|c}{XPOS correction} \\
Original & & Correction & Count & Ratio & Orig & & Correction & Count & Ratio\\\hline 
ADV &$\rightarrow$& NOUN & 3607 & 0.0636 &   XR+XSA+ETM &$\rightarrow$& NNG+XSA+ETM & 321 & 0.0056 \\
NOUN &$\rightarrow$& PROPN & 2654 & 0.0468& XR+XSA+EC &$\rightarrow$& NNG+XSA+EC & 248 & 0.0043 \\
VERB &$\rightarrow$& ADJ & 1065 & 0.0187& NNG+JKS &$\rightarrow$& NNG+JKC & 51 & 0.0008 \\
ADV &$\rightarrow$& PROPN & 404 & 0.0071& XR &$\rightarrow$& NNG & 44 & 0.0007 \\
ADV &$\rightarrow$& ADJ & 268 & 0.0047& MAG &$\rightarrow$& MAJ & 44 & 0.0007 \\ \hline 
\end{tabular}}
\caption{Top-5 UPOS and XPOS conversion results} \label{pos-conversion-results}
\end{table*}

\subsection{Integration with UniDive}

\paragraph{Morphosyntactic feature extraction and rule-based annotation}
Each word in the Korean UniDive dataset is analyzed at the morpheme level, allowing for the extraction of morphosyntactic features based on both individual morphemes and their corresponding POS tags. {Feature assignment is governed by an extensive set of linguistic rules grounded in Korean's agglutinative structure}, including suffix-based inflection, grammatical function markers, and verb-final {particles}. By incorporating fine-grained segmentation and systematic feature extraction, we accurately capture case markers, tense and aspect distinctions, and honorific expressions, ensuring the preservation of essential morphosyntactic information. This rule-based refinement enhances the alignment between morphological and syntactic annotations, contributing to the dataset's overall consistency and linguistic accuracy. Table~\ref{ms-table} explicitly summarizes the extraction rules applied in this process.

\paragraph{Functional words}

\jp{As part of our UniDive adaptation for Korean, we manually selected a set of potential functional words from the Sejong corpus, focusing on high-frequency adverbs ({MAG}) and determiners ({MM}). These functional categories often contribute to sentence structure and meaning without introducing new content, making them particularly relevant for syntactic and typological analysis. For adverbs, we selected frequently occurring items such as {더} \textit{deo} (`more'), {또} \textit{tto} (`again'), and {다시} \textit{dasi} (`again'), which often express degree, repetition, or negation. For determiners, we included high-frequency words like {그} \textit{geu} (`that'), {이} \textit{i} (`this'), and {한} \textit{han} (`one'), which are commonly used in reference and quantification. These selections are intended to represent functionally salient elements within Korean, facilitating cross-linguistic comparisons in the UniDive framework.}

\paragraph{Conjunctive verbal endings}
\jp{For verbs—particularly those that do not carry morphosyntactic features—we follow the UniDive convention of adding a transcription of the verbal ending.
For example, a verb with the conjunctive ending {가서} \textit{ga-seo} (`go + \textit{seo}') can be annotated with a feature such as \texttt{Case=seo} to reflect the surface form of the ending.
Many conjunctive verbal endings in Korean, however, are semantically rich and encode relations such as causality, contrast, sequence, or condition. Given the complexity and variability of these meanings, we leave a more fine-grained semantic classification of conjunctive endings as future work.}

\subsection{Experimental setup}

To evaluate the {effectiveness} of incorporating UniMorph-derived morphosyntactic features, we conduct dependency parsing experiments using the Korean UniDive dataset. Our objective is to determine how enriched morphosyntactic representations influence parsing performance across different models.
In this study, we train two types of dependency parsing models to examine the effects of morphosyntactic features on syntactic analysis. The first is an {encoder-only} model using \texttt{UDPipe 1} \citep{straka-strakova:2017:CoNLL}, which serves as a baseline for conventional deterministic parsing \jp{and makes use of morphosyntactic features during training.}
The second is a {decoder-only} model, implemented using \texttt{Bllossom-3B} \citep{choi-etal-2024-optimizing}, a fine-tuned variant of the Llama-family model, trained via instruction tuning. These models allow for a comparative assessment of how structured linguistic features contribute to parsing accuracy.
The training procedure follows the standard data splits provided by the Korean GSD treebank, ensuring a controlled experimental setup. Each model is trained both with and without UniMorph-derived morphosyntactic features, allowing us to {compare their performances}. The {encoder-only} approach follows a more traditional dependency parsing methodology, whereas the {decoder-only} model reconstructs syntactic structures generatively, utilizing instruction tuning. It is important to clarify that our aim is not to compare the absolute performance of \texttt{UDPipe 1} and \texttt{Bllossom-3B}, as these are fundamentally different {models}. Rather, the goal is to investigate how morphosyntactic enhancements influence each model type, highlighting the impact of linguistic features on dependency parsing.

\subsection{Results}

Table~\ref{parsing-results} presents the results of our dependency parsing experiments. The findings indicate that integrating morphosyntactic features consistently improves parsing accuracy across both {encoder-only and decoder-only} models.
The incorporation of morphosyntactic features leads to significant gains in parsing performance. The {encoder-only} \texttt{UDPipe} model, which initially struggles with Korean's complex morphology, exhibits the most pronounced improvements, with an increase of 14.09 LAS points. This enhancement demonstrates that explicit morphological information plays a crucial role in refining syntactic structures, particularly in feature-driven dependency relations. For the {decoder-only} \texttt{Bllossom-3B} model, improvements are more modest but still notable, with LAS increasing by 2.6 points. Since large-scale generative models inherently capture some morphosyntactic properties during pretraining, the additional morphosyntactic annotations serve as complementary refinements rather than primary determinants of performance.

\begin{table}[!ht]
\centering
\footnotesize{
\begin{tabular}{rr cc}
\hline
&  & UAS & LAS \\
\hline
\texttt{udpipe 1}& Without \textsc{ms} features   &  61.05 &  50.24 \\
& With \textsc{ms} features &  71.41 &  64.33 \\ 
\hdashline
\texttt{bllossom-3B} & Without \textsc{ms} features   &  88.30 &  84.37 \\
& With \textsc{ms} features &  \textbf{89.16} &  \textbf{86.97} \\
\hline
\end{tabular}
}
\caption{Dependency parsing results with and without morphosyntactic features}
\label{parsing-results}
\end{table}

\section{Discussion}

\paragraph{Alignment with Korean linguistic theories}
The integration of morphosyntactic features into dependency parsing aligns with Korean linguistic {theories}, particularly in case marking, argument structure, and predicate composition. As an agglutinative language, Korean relies on suffix-based morphology to encode grammatical relations, making explicit morphological annotations essential for syntactic disambiguation. Traditional linguistic analyses emphasize postpositional case markers in distinguishing subjects, objects, and adjuncts, {which directly benefits} parsing models when these features are incorporated.

Our findings reflect theoretical accounts of Korean syntax, where syntactic roles are determined by morphological markers rather than fixed word order. Explicit case distinctions mitigate subject-object confusion, especially in non-canonical SOV structures. Additionally, refined verbal morphology representation, including tense, aspect, and modality, {enhances the analysis of predicate-argument structures and captures} Korean's complex predicates involving auxiliary verbs and honorifics.

Our approach also differentiates {functional} and {lexical} morphemes, ensuring more accurate dependency relations. {Functional} morphemes like case markers and sentence-final endings often cause misclassification in models relying on surface forms alone. By incorporating detailed morphosyntactic features, we {obtain} a linguistically informed representation that improves parsing accuracy in feature-driven syntactic relations.

\paragraph{Cross-linguistic applicability}
To assess the broader applicability of our approach beyond Korean, we conduct additional experiments on the UD Turkish Penn Treebank, a dataset that similarly incorporates rich morphosyntactic annotations. Turkish, like Korean, is an agglutinative language with extensive suffixation and complex morphosyntactic dependencies, making it an ideal test case for evaluating the effectiveness of our method in a cross-linguistic setting. 

The experimental setup remains identical to our Korean experiments, ensuring a controlled comparison of the impact of morphosyntactic features on dependency parsing. Preliminary results indicate that the integration of morphosyntactic annotations leads to improvements in parsing accuracy for Turkish, {a tendency that aligns with findings from experiments on the Korean GSD treebank}. These findings suggest that our approach is not only effective for Korean but also extends to other morphologically rich languages, reinforcing the importance of explicit morphosyntactic representations in dependency parsing across diverse linguistic typologies.

\paragraph{Future directions}
UniMorph provides word-level morphological predictions, independently assigning features without considering sentence context. In contrast, our experiments assumed gold-standard morphosyntactic features, directly integrating them into dependency parsing.
As future work, we aim to bridge this gap by developing a sentence-level feature prediction pipeline. This involves first predicting word-level features using UniMorph-style annotation methods, then contextually refining them based on syntactic dependencies. By propagating features across dependency arcs, we can resolve ambiguities and enhance their alignment with sentence structure.
Extending word-level UniMorph predictions to sentence-level applications will improve {parsing} accuracy, particularly for morphologically rich languages. Future work will focus on refining feature propagation and interaction within dependency parsing models, ensuring a seamless integration of morphosyntactic annotations in structured syntactic analysis.

\section{Conclusion}

In this work, we developed the Korean UniDive dataset, integrating UniMorph-derived morphosyntactic features with Universal Dependencies to enhance Korean dependency parsing. Our dataset systematically aligns morphological and syntactic annotations, addressing inconsistencies in existing resources and improving linguistic representation. We demonstrate that explicit morphosyntactic annotations significantly enhance {encoder-only} models, while providing complementary refinements for {decoder-only} generative models.

\section*{Limitations}
While the integration of UniMorph and Universal Dependencies under the UniDive framework presents a promising approach to representing Korean morphosyntactic structures, several limitations remain. These challenges highlight areas for further refinement and potential extensions to enhance the framework's applicability.
One of the inherent difficulties in modeling Korean within UD is the word segmentation issue. Unlike languages with clear word boundaries, Korean's agglutinative structure means that functional morphemes (e.g., case markers, verb endings) are tightly linked to content words. While UniDive addresses this by encoding functional morphemes as morphological features rather than separate syntactic nodes, boundary inconsistencies persist when transitioning between different annotation schemes.

While our dataset enhances the representation of core morphosyntactic structures, certain non-canonical constructions remain challenging. These include: (i) elliptical structures, where Korean frequently omits subjects or objects, requiring models to infer missing elements from context \citep{han-etal-2020-null}; (ii) periphrastic constructions, where multi-word verb phrases, such as 먹어 보고 싶다 \textit{meog-eo bogo sipda} (`want to try eating'), introduce structural complexity that UniDive does not fully capture \citep{chung-1998-arugment}; and (iii) multi-function morphemes, where certain morphemes serve multiple syntactic roles depending on context (e.g., -고 \textit{go}, which can indicate either coordination or sequencing) \citep{park-kim-2023-role}.
These limitations suggest that a more fine-grained approach to Korean morphosyntactic variation may be necessary for future work.

\section*{Acknowledgments}
We would like to thank Omer Goldman and the UniDive Consortium for their valuable support and collaboration in building the Korean UniDive dataset.

% \bibliography{references}

\appendix
% \onecolumn

\section{Licenses}
Our Korean UniDive dataset, derived from the Universal Dependencies (UD) Korean GSD Treebank, is distributed under the Creative Commons Attribution-ShareAlike 4.0 International License (CC BY-SA 4.0). This license allows users to share (copy and redistribute the material in any medium or format) and adapt (remix, transform, and build upon the material for any purpose, including commercial use), provided that proper attribution is given to the original dataset and that any derivative works are released under the same license.

\section{Technical Details on Experiments}
\label{appendix:exp-settings}

\paragraph{Optimization and training strategy of LLM} 
Table~\ref{appendix-hyperparameter-settings} provides an overview of the hyperparameter configuration. We train the model using AdamW-8bit, a memory-friendly derivative of AdamW, with a weight decay of 0.01 to mitigate overfitting. The initial learning rate is set to 5e-5 and is gradually lowered according to a cosine schedule. Training is conducted with a batch size of 1, and gradient accumulation is performed for 4 steps to attain an effective batch size of 4.

\paragraph{GPUs used}
We used a single NVIDIA A6000 GPU for both training and evaluation. Training took approximately 47 minutes per epoch, and the model was trained for a total of 10 epochs.

\begin{table}[!th]
    \centering
    {\footnotesize
    \begin{tabular}{ll}
        \hline
        {Hyperparameter}      & {Value}  \\
        \hline
        D-type                      & bfloat16    \\
        Learnign-rate               & 5e-5        \\
        Warm-up ratio               & 0.03        \\
        Learning-rate scheduler     & cosine      \\
        Optimizer                   & AdamW-8bit  \\
        Weight decay                & 0.01     \\
        Batch size                  & 1        \\
        Gradient accumulation steps & 16       \\
        Training epochs             & 1        \\
        \hline
    \end{tabular}
    }
    \caption{Hyperparameter settings used for training}
    \label{appendix-hyperparameter-settings}
\end{table}

\end{document}